\documentclass{article} % For LaTeX2e
\usepackage{iclr2021_conference,times}

\usepackage{amsmath,amsfonts,bm}

\def\figref#1{figure~\ref{#1}}
\def\secref#1{section~\ref{#1}}
\def\eqref#1{equation~\ref{#1}}
\def\1{\bm{1}}

\def\vp{{\bm{p}}}

\def\vr{{\bm{r}}}

\def\vu{{\bm{u}}}

\def\vx{{\bm{x}}}
\def\vy{{\bm{y}}}

\def\mX{{\bm{X}}}

\DeclareMathAlphabet{\mathsfit}{\encodingdefault}{\sfdefault}{m}{sl}
\SetMathAlphabet{\mathsfit}{bold}{\encodingdefault}{\sfdefault}{bx}{n}

\def\gA{{\mathcal{A}}}
\def\gB{{\mathcal{B}}}

\def\gE{{\mathcal{E}}}

\def\gK{{\mathcal{K}}}
\def\gL{{\mathcal{L}}}

\def\gR{{\mathcal{R}}}

\def\gT{{\mathcal{T}}}
\def\gU{{\mathcal{U}}}

\def\gY{{\mathcal{Y}}}
\def\gZ{{\mathcal{Z}}}

\def\sQ{{\mathbb{Q}}}
\def\sR{{\mathbb{R}}}

\def\sX{{\mathbb{X}}}
\def\sY{{\mathbb{Y}}}

\def\emH{{H}}

\DeclareMathOperator*{\argmax}{arg\,max}

\DeclareMathOperator*{\logadd}{logadd}
\newcommand{\ctcblank}{\texttt{\textless b\textgreater}}
\newcommand{\start}{\small \texttt{\textless s\textgreater}}
\newcommand{\wordsep}[1]{\underline{\hspace{2mm}}#1}

\usepackage{hyperref}
\usepackage{booktabs}
\usepackage{graphicx}
\usepackage{listings}
\usepackage{multirow}
\usepackage{pgfplots}
\usepackage{subcaption}
\usepackage{wrapfig}
\usepackage{url}
\usepackage{tikz}
\usetikzlibrary{arrows,automata,positioning}

\pgfplotsset{
  cycle multiindex* list={
    black white \nextlist linestyles \nextlist mark list },
  /tikz/mark size=1.5pt
}

\title{Differentiable Weighted Finite-State \\
Transducers}

\author{Awni Hannun, Vineel Pratap, Jacob Kahn \& Wei-Ning Hsu \\
Facebook AI Research \\
\texttt{\{awni,vineelkpratap,jacobkahn,wnhsu\}@fb.com}}

\iclrfinalcopy % Uncomment for camera-ready version, but NOT for submission.
\begin{document}

\maketitle

\begin{abstract}
We introduce a framework for automatic differentiation with weighted
finite-state transducers (WFSTs) allowing them to be used dynamically at
training time. Through the separation of graphs from operations on graphs,
this framework enables the exploration of new structured loss functions which
in turn eases the encoding of prior knowledge into learning algorithms. We
show how the framework can combine pruning and back-off in transition models
with various sequence-level loss functions. We also show how to learn over
the latent decomposition of phrases into word pieces. Finally, to
demonstrate that WFSTs can be used in the interior of a deep neural network,
we propose a convolutional WFST layer which maps lower-level representations
to higher-level representations and can be used as a drop-in replacement for a
traditional convolution. We validate these algorithms with experiments in
handwriting recognition and speech recognition.
\end{abstract}

\section{Introduction}
\label{sec:intro}
Weighted finite-state transducers (WFSTs) are a commonly used tool in speech
and language processing~\citep{knight2009applications, mohri2002weighted}. They
are most frequently used to combine predictions from multiple already trained
models. In speech recognition, for example, WFSTs are used to combine
constraints from an acoustic-to-phoneme model, a lexicon mapping words to
pronunciations, and a word-level language model. However, combining separately
learned models using WFSTs only at inference time has several drawbacks,
including the well-known problems of \emph{exposure
bias}~\citep{ranzato2015sequence} and \emph{label bias}~\citep{bottou91,
lafferty2001}.

Given that gradients may be computed for most WFST operations,
using them only at the inference stage of a learning system is not a hard
limitation. We speculate that this limitation is primarily due to practical
considerations. Historically, hardware has not been sufficiently performant to
make training with WFSTs tractable. Also, no implementation exists with the
required operations which supports automatic differentiation in a high-level
yet efficient manner.

We develop a framework for automatic differentiation through operations on
WFSTs. We show the utility of this framework by leveraging it to design
and experiment with existing and novel learning algorithms. Automata are a more
convenient structure than tensors to encode prior knowledge into a learning
algorithm. However, not training with them limits the extent to which this
prior knowledge can be incorporated in a useful manner. A framework for
differentiable WFSTs allows the model to jointly learn from training data as
well as prior knowledge encoded in WFSTs. This enables the learning algorithm
to incorporate such knowledge in the best possible way.

Use of WFSTs conveniently decomposes operations from data ({\emph i.e.}
graphs). For example, rather than hand-coding sequence-level loss functions
such as Connectionist Temporal Classification (CTC)~\citep{graves2006} or the
Automatic Segmentation Criterion (ASG)~\citep{collobert2016wav2letter}, we may
specify the core assumptions of the criteria in graphs and compute the
resulting loss with graph operations. This facilitates exploration in the space
of such structured loss functions.

We show the utility of the differentiable WFST framework by designing and
testing several algorithms. For example, bi-gram transitions may be added to
CTC with a transition WFST. We scale transitions to large token set
sizes by encoding pruning and back-off in the transition graph.

Word pieces are commonly used as the output of speech recognition and machine
translation models~\citep{chiu2018state, sennrich2016}. The word piece
decomposition for a word is learned with a task-independent model. Instead,
we use WFSTs to marginalize over the latent word piece decomposition at
training time. This lets the model learn decompositions salient to the task at
hand.

Finally, we show that WFSTs may be used as layers themselves intermixed with
tensor-based layers. We propose a convolutional WFST layer which maps
lower-level representations to higher-level representations. The WFST
convolution can be trained with the rest of the model and results in improved
accuracy with fewer parameters and operations as compared to a traditional
convolution.

In summary, our contributions are:
\begin{itemize}
\item A framework for automatic differentiation with WFSTs. The framework
  supports both C++ and Python front-ends and is available at
    \url{https://www.anonymized.com}.

\item We show that the framework may be used to express both existing
  sequence-level loss functions and to design novel sequence-level loss
  functions.

\item We propose a convolutional WFST layer which can be used in the interior
  of a deep neural network to map lower-level representations to higher-level
  representations.

\item We demonstrate the effectiveness of using WFSTs in the manners described
  above with experiments in automatic speech and handwriting recognition.

\end{itemize}

\begin{figure}
\centering
\begin{lstlisting}[language=Python, frame=single,
    basicstyle=\ttfamily\scriptsize,
    keywordstyle=\bfseries\color{magenta!70!black},
    commentstyle=\bfseries\color{green!50!black},
    otherkeywords={True, False},
    xleftmargin=20pt,
    xrightmargin=20pt,
    ]
from gtn import *

def ASG(emissions, transitions, target):
  # Compute constrained and normalization graphs:
  A = intersect(intersect(target, transitions), emissions)
  Z = intersect(transitions, emissions)

  # Forward both graphs:
  A_score = forward_score(A)
  Z_score = forward_score(Z)

  # Compute loss:
  loss = negate(subtract(A_score, Z_score))

  # Clear previous gradients:
  emissions.zero_grad()
  transitions.zero_grad()

  # Compute gradients:
  backward(loss, retain_graph=False)
  return loss.item(), emissions.grad(), transitions.grad()
}
\end{lstlisting}
\caption{An example using the Python front-end of \texttt{gtn} to compute the
  ASG loss function and gradients. The inputs to the \texttt{ASG} function are
  all \texttt{gtn.Graph} objects.}
\label{fig:asg_python}
\end{figure}

\section{Related Work}
\label{sec:related}

A wealth of prior work exists using weighted finite-state automata in speech
recognition, natural language processing, optical character recognition, and
other applications~\citep{breuel2008ocropus, knight2009applications,
mohri1997finite, mohri2008speech, pereira1994weighted}. However, the use of
WFSTs is limited mostly to the inference stage of a predictive system. For
example, Kaldi, a commonly used toolkit for automatic speech recognition, uses
WFSTs extensively, but in most cases for inference or to estimate the
parameters of shallow models~\citep{povey2011kaldi}. In some cases, WFSTs are
used statically to incorporate fixed lattices in discriminative sequence
criteria~\citep{vesely2013sequence}.

Implementations of sequence criteria in end-to-end style training are typically
hand-crafted with careful consideration for speed~\citep{amodei2016deep,
collobert2019fully, povey2016purely}. The use of hand-crafted implementations
reduces flexibility which limits research. In some cases, such as the fully
differentiable beam search of~\citet{collobert2019fully}, achieving the
necessary computational efficiency with a WFST-based implementation may not yet
be tractable. However, as a first step, we show that in many common cases we
can have the expressiveness afforded by the differentiable WFST framework
without paying an unacceptable penalty in execution time.

Related to this work, and inspiring the name of our framework, are the graph
transformer networks of~\citet{bottou97}. Generalized graph
transducers~\citep{bottou96} are more expressive than WFSTs, allowing
arbitrary data as edge labels. When composing these graphs, one defines an edge
matching function and a ``transformer'' to construct the resulting structure.
While not this general, differentiable WFSTs nevertheless allow for a vast
design space of interesting algorithms, and perhaps make a more pragmatic
trade-off between flexibility and efficiency.

Highly efficient libraries for operations on WFSTs exist, notably OpenFST and
its predecessor FSM~\citep{allauzen2007openfst, mohri2000design}. We take
inspiration from OpenFst in the interface and implementation of many of our
functions. However, the design implications of operating on WFSTs with
automatic differentiation are quite different than those of the use cases
OpenFST has been optimized for. We also draw inspiration from libraries for
automatic differentiation and deep learning~\citep{collobert2011torch7,
paszke2019pytorch, pratap2019wav2letter++, tokui2015chainer}.

Some of the algorithms we propose, with the goal of demonstrating the utility
of the differentiable WFST library, are inspired by prior work. Prior work has
explored pruning the set of allowed alignments with CTC, and in particular
limiting the spacing between output tokens~\citep{liu2018connectionist}.
Learning $n$-gram word decompositions with both
differentiable~\citep{liu2017gram} and
non-differentiable~\citep{chan2016latent} loss functions has also been
explored.

\section{Differentiable Weighted Finite-State Transducers}
\label{sec:wfsts}

A weighted finite-state acceptor $\gA$ is a $6$-tuple consisting of an alphabet
$\Sigma$, a set of states $\sQ$, start states $\sQ_s$, accepting states
$\sQ_a$, a transition function $\pi(q, p)$ which maps elements of $\sQ \times
\Sigma$ to elements of $\sQ$, and a weight function $\omega(q, p)$ which maps
elements of $\sQ \times \Sigma$ to $\sR$. A weighted finite-state transducer
$\gT$ is a $7$-tuple which augments an acceptor with an output alphabet
$\Delta$. The transition function $\pi(q, p, r)$ and weight function $\omega(q,
p, r)$ map elements of $\sQ \times \Sigma \times \Delta$ to elements of $\sQ$
and $\sR$ respectively. In other words, each edge of a transducer connects two
states and has an input label $p$, an output label $r$ and a weight $w \in
\sR$.

We denote an input by $\vp = [p_1, \ldots, p_T]$ where each $p_i \in \Sigma$.
An acceptor $\gA$ \emph{accepts} the input $\vp$ if there exists a sequence of
states $q_{i+1} = \pi(q_i, p_i) \in \sQ$ such that $q_1 \in \sQ_s$ and $q_{T+1}
\in \sQ_a$. With a slight abuse of notation, we let $\vp \in \gA$ denote that
$\gA$ accepts $\vp$. The score of $\vp$ is given by $s(\vp) = \sum_{i=1}^T
\omega(q_i, p_i)$. Let $\vr = [r_1, \dots, r_T]$ be a path with $r_i \in
\Delta$. A transducer $\gT$ \emph{transduces} the input $\vp$ to the output
$\vr$ (\emph{i.e.} $(\vp, \vr) \in \gT$) if there exists a sequence of states
$q_{i+1} = \pi(q_i, p_i, r_i) \in \sQ$ such that $q_1 \in \sQ_s$ and $q_{T+1}
\in \sQ_a$. The score of the pair $(\vp, \vr)$ is given by $s(\vp, \vr) =
\sum_{i=1}^T \omega(q_i, p_i, r_i)$.

We restrict operations to the log and tropical semirings. In both cases
the accumulation of weights along a path is with addition. In the log semiring
the accumulation over path scores is with log-sum-exp which we denote by
$\logadd_{i} s_i = \log \sum_{i} e^{s_i}$. In the tropical semiring, the
accumulation over path scores is with the max, $\max_{i} s_i$. We allow
$\epsilon$ transitions in both acceptors and transducers. An $\epsilon$ input
(or output) on an edge means the edge can be traversed without consuming
an input (or emitting an output). This allows inputs to map to outputs of
differing length.

\subsection{Operations}
\label{sec:operations}

We briefly describe a subset of the most important operations implemented in
our differentiable WFST framework. For a more detailed discussion of operations
on WFSTs see \emph{e.g.} ~\citet{mohri2009weighted}.

{\bf Intersection} We denote by $\gB = \gA_1 \circ \gA_2$ the intersection of
two acceptors $\gA_1$ and $\gA_2$. The intersected graph $\gB$ contains all
paths which are accepted by both inputs. The score of any path in $\gB$ is the
sum of the scores of the two corresponding paths in $\gA_1$ and $\gA_2$.

{\bf Composition} The same symbol denotes the composition of two transducers
$\gU = \gT_1 \circ \gT_2$. If $\gT_1$ transduces $\vp$ to $\vu$ and $\gT_2$
transduces $\vu$ to $\vr$ then $\gU$ transduces $\vp$ to $\vr$. As in
intersection, the score of the path in the composed graph is the sum of the
path scores from the input graphs.

{\bf Forward and Viterbi Score} The forward score of $\gT$ is $\logadd_{(\vp,
\vr) \in \gA} s(\vp, \vr)$. Similarly, the Viterbi score of $\gT$ is
$\max_{(\vp, \vr) \in \gT} s(\vp, \vr)$.

{\bf Viterbi Path} The Viterbi path of $\gT$ is given by $\argmax_{(\vp, \vr)
\in \gT} s(\vp, \vr)$.

The forward score, Viterbi score, and the Viterbi path for an acceptor $\gA$
are defined in the same way using just $\vp$. In a directed acyclic graph these
operations can be computed in time linear in the size of the graph with the
well-known forward and Viterbi algorithms~\citep{jurafsky2000speech,
rabiner1989tutorial}.

We support standard rational operations including the union ($\gA + \gB$),
concatenation ($\gA\gB$), and the Kleene closure ($\gA^*$) of WFSTs. To enable
automatic differentiation through complete computations, we support slightly
non-standard operations. We allow for the negation of all the arcs weights in a
graph and the addition or subtraction of the arc weights of two identically
structured graphs.

\subsection{Automatic Differentiation}

For the sake of automatic differentiation, every operation in
\secref{sec:operations} accepts as inputs one or more graphs and returns a
graph as output. For example, the forward score returns a ``scalar" graph with
a single arc between a start state and accepting state, with the score
as the arc weight. The Viterbi path outputs the linear graph representing the
best path $\vp^*$ with $p^*_i$ as the arc labels. Jacobians of the output graph
weights with respect to the weights of the input graphs may be computed for all
of the operations described so far. Hence, the chain rule may be used to
compute Jacobians or gradients of outputs with respect to the inputs of
compositions of graph operations.

The WFST framework implements reverse-mode automatic differentiation following
existing deep learning frameworks~\citep{pratap2019wav2letter++,
paszke2019pytorch}. Figure~\ref{fig:asg_python} gives an example Python
implementation which computes gradients for the ASG criterion.  In order to
enable second-order differentiation, among other design considerations, the
gradient with respect to a graph is also a graph. This adds little additional
overhead as the two graphs share the underlying graph topology. The weights of
the gradient graph are the gradients with respect to the weights of the graph.

\begin{figure}
    \centering
    \begin{subfigure}[b]{0.32\textwidth}
      \centering
      \includegraphics[width=0.5\linewidth]{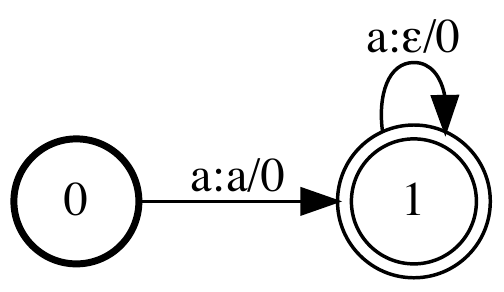}
      \caption{Token graph $\gT_a$}
      \label{fig:asg_token_a}
    \end{subfigure}
    \begin{subfigure}[b]{0.32\textwidth}
        \centering
        \includegraphics[width=0.9\linewidth]{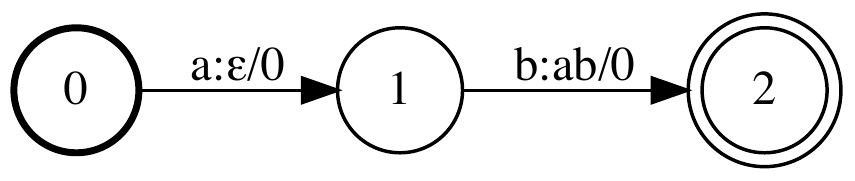}
        \caption{Label graph $\gY$}
        \label{fig:asg_label}
    \end{subfigure}
    \begin{subfigure}[b]{0.32\textwidth}
        \centering
        \includegraphics[width=0.9\linewidth]{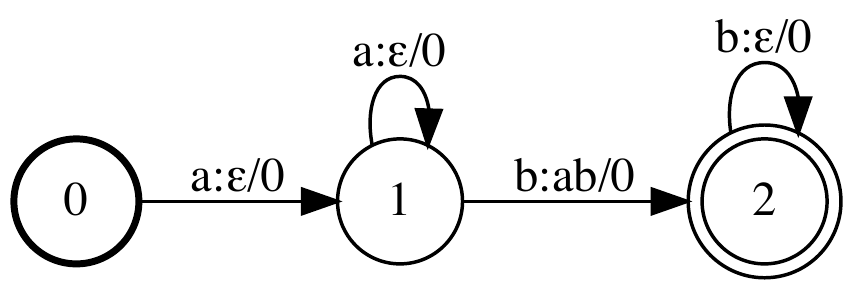}
        \caption{Alignment graph $\gT \circ \gY$}
        \label{fig:asg_alignments}
    \end{subfigure}
    \par\medskip
    \begin{subfigure}[b]{0.45\textwidth}
        \centering
        \includegraphics[width=1.0\linewidth]{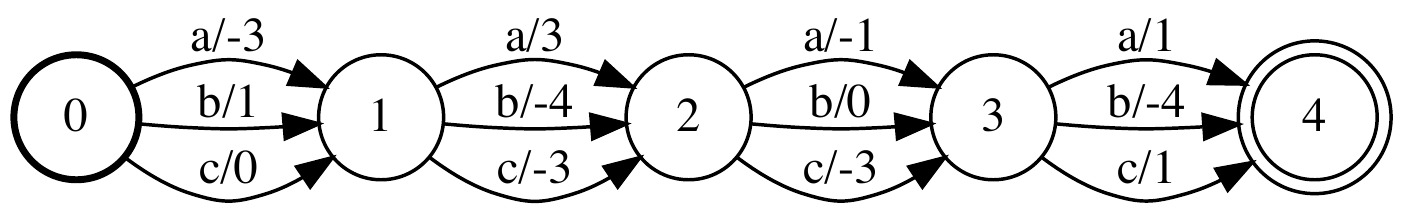}
        \caption{Emissions graph $\gE_{\mX}$}
        \label{fig:asg_emissions}
    \end{subfigure}
    \hspace{2mm}
    \begin{subfigure}[b]{0.45\textwidth}
        \centering
        \includegraphics[width=1.0\linewidth]{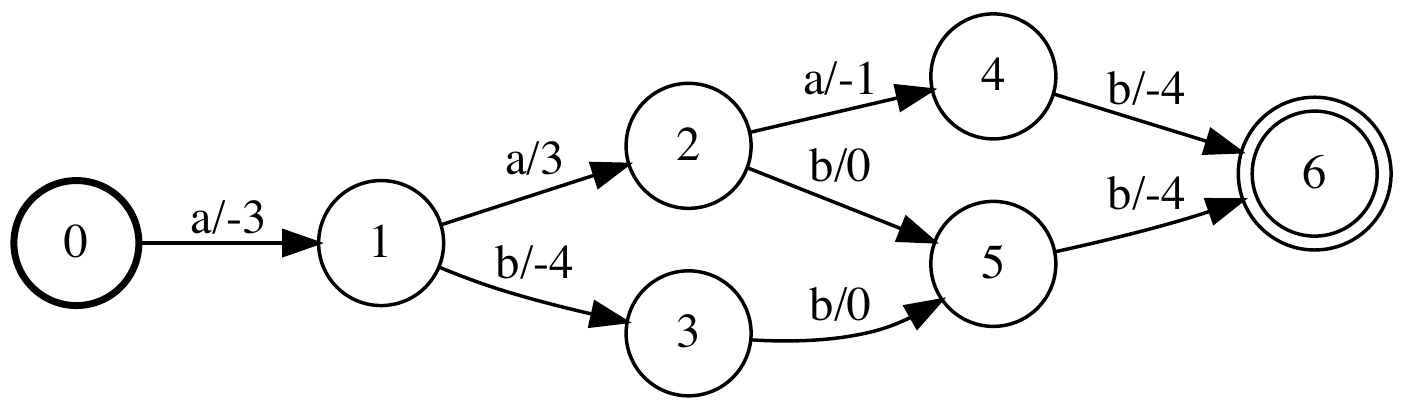}
        \caption{Constrained graph $\gA = \gT \circ \gY \circ \gE_{\mX}$}
        \label{fig:asg_constrained}
    \end{subfigure}
    \caption{The graphs used to construct the ASG sequence criterion. The arc
      label ``p:r/w" denotes an input label $p$, an output label $r$ and weight
      $w$. Graphs with just ``p/w'' are acceptors.}
    \label{fig:asg_graphs}
\end{figure}

\section{Learning Algorithms}
\label{sec:algorithms}

Let $\mX = [\vx_1, \ldots \vx_T] \in \sX$ be a sequence of observations and
$\vy = [y_1, \ldots, y_U] \in \sY$ a label. A sequence-level objective for the
$(\mX, \vy)$ pair can be specified with two weighted automata:
\begin{equation}
    \label{eq:general_sequence_criterion}
    \log p(\vy \mid \mX) = \sum_{\vp \in \gA_{\mX, \vy}} \logadd s(\vp) - \sum_{\vp \in \gZ_{\mX}} \logadd s(\vp).
\end{equation}
The target constrained graph, $\gA_{\mX, \vy}$, is constructed from the
observation and target pair and the normalization graph, $\gZ_{\mX}$, is
constructed only from the observation. We require $\gA_{\mX, \vy} \subseteq \gZ_\mX$.

The key ingredients to make \eqref{eq:general_sequence_criterion} operational
are 1) the structure of $\gA$ and $\gZ$ and 2) the source of the arc weights.
The graph structures can be specified through functions on graphs. The initial
arc weights come from the data itself or arbitrary learning algorithms (deep
networks, $n$-gram language models, \emph{etc.}) capable of providing useful
scores on elements of $\sX$ and $\sY$.

\begin{figure}
\centering
\begin{subfigure}[b]{.2\textwidth}
  \centering
  \includegraphics[width=0.3\linewidth]{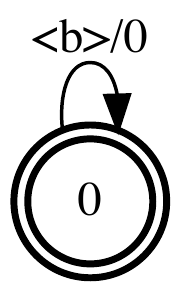}
  \caption{Blank token graph.}
  \label{fig:ctc_blank}
\end{subfigure}
\begin{subfigure}[b]{.7\textwidth}
  \centering
  \includegraphics[width=0.9\linewidth]{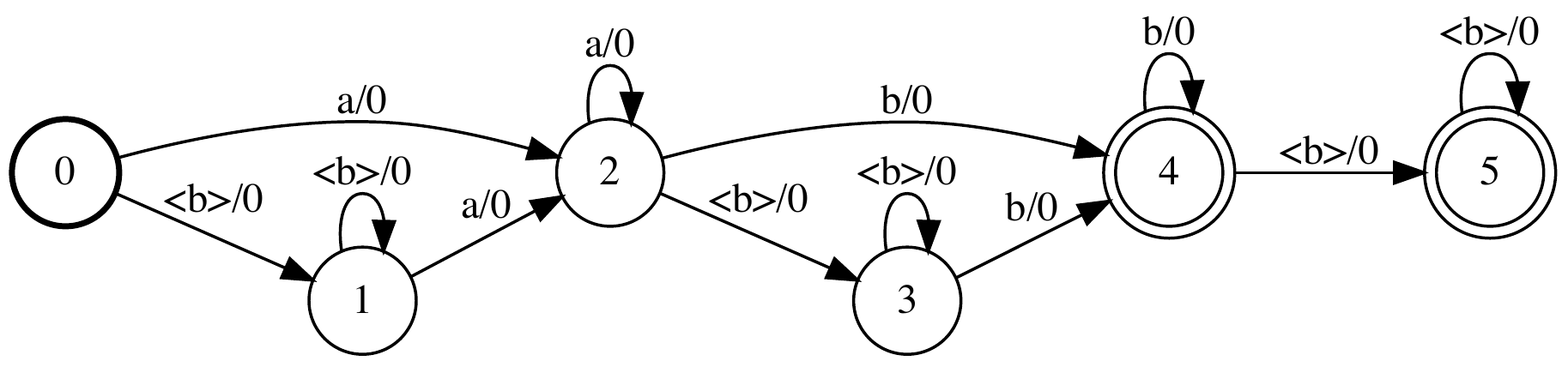}
  \caption{Alignment graph ($\gT \circ \gY$).}
  \label{fig:ctc_alignments}
\end{subfigure}
\caption{The primary difference between ASG and CTC is the inclusion of the
  blank token graph (a) which allows for optional transitions on \ctcblank\, and
  results in the CTC alignment graph (b).}
\end{figure}

The graphs $\gA$ and $\gZ$ used by the ASG loss function can be constructed
from operations on simpler graphs. The ASG criterion assumes that emitting a
new token requires consuming the token at least once, as in the graph, $\gT_a$,
of \figref{fig:asg_token_a}. The full token graph is the closure of the union
of the individual token graphs. Assuming an alphabet of $\{$a, b, c$\}$ gives
$\gT = (\gT_a + \gT_b + \gT_c)^*$. Composing $\gT$ with the label graph $\gY$
(fig.~\ref{fig:asg_label}) of ``ab'' results in the alignment graph
(fig.~\ref{fig:asg_alignments}). The alignment graph specifies the allowed
correspondences between arbitrary length paths and the desired label.
Intersecting the alignment graph with the emissions graph $\gE_\mX$
(fig.~\ref{fig:asg_emissions}) results in the constrained graph $\gA$
(fig.~\ref{fig:asg_constrained}). The arc weights of $\gE_\mX$ depend on $\mX$
and can be the output of a deep neural network. Here $\gZ$ is simply the
emissions graph. We may add bigram transitions by including a bigram
transition graph, $\gB$, as in~\figref{fig:bigram_transitions}. In summary, the
graph operations are:
\begin{equation}
    \label{eq:asg_from_graphs}
    \gA_{\textrm{ASG}} = \gE \circ (\gB \circ ((\gT_1 + \ldots + \gT_C)^* \circ \gY)) \quad \textrm{and} \quad
    \gZ_{\textrm{ASG}} = \gE \circ \gB.
\end{equation}
The primary difference between ASG and CTC is the inclusion of a blank token,
\ctcblank, represented by the graph in \figref{fig:ctc_blank}. Constructing CTC
amounts to including the blank token graph when constructing the full token
graph $\gT$. The intersection $\gT \circ \gY$ then results in the CTC alignment
graph (fig.~\ref{fig:ctc_alignments}). Note, this version of CTC does not force
transitions on \ctcblank\, between repeats tokens. This requires remembering the
previous state and hence is more involved (see
Appendix~\ref{sec_apx:ctc_repeats} for details).

A benefit of constructing sequence-level criteria by composing operations on
simpler graphs is the access to a large design space of loss functions with
which we can encode useful priors. For example we could construct a ``spike''
CTC, a ``duration-limited'' CTC, or an ``equally spaced'' CTC by substituting
the appropriate token graphs into \eqref{eq:asg_from_graphs} (see
Appendix~\ref{sec_apx:ctc_priors} for details).

\subsection{Transitions}

\begin{figure}
\centering
\begin{subfigure}[b]{0.44\textwidth}
  \centering
  \resizebox{1.0\textwidth}{!}{
    \begin{tikzpicture}
 [>=latex,
  double distance=2pt,
  draw,-latex,
  line width=1pt,
  initial text={},
  shorten >=1pt,
  node distance=3.5cm,
  auto,
  every state/.style={draw=black, thick}
 ]
  \node[state, initial, accepting] (s) {\start};
  \node[state, accepting, right of=s] (b) {b};
  \node[state, accepting, above right of=b] (a) {a};
  \node[state, accepting, below right of=b] (c) {c};
  \draw (s) edge[bend left] node {a/$w$(\start, a)} (a);
  \draw (s) edge node {b/$w$(\start, b)} (b);
  \draw (s) edge[bend right]  node [midway, left=15pt]{c/$w$(\start, c)} (c);
  \draw (a) edge[loop above] node [left=5pt] {a/$w$(a, a)} (a);
  \draw (a) edge[bend right=25pt] node [near start, left=5pt] {b/$w$(a, b)} (b);
  \draw (a) edge[bend left=5pt] node {c/$w$(a, c)} (c);
  \draw (b) edge[bend right=40pt] node [near end, left=0pt] {a/$w$(b, a)} (a);
  \draw (b) edge[loop above] node [left=5pt] {b/$w$(b, b)} (b);
  \draw (b) edge[bend right=40pt] node [near start, left=0pt] {c/$w$(b, c)} (c);
  \draw (c) edge[bend right=70pt] node [near start, right] {a/$w$(c, a)} (a);
  \draw (c) edge[bend right=30pt] node [near start, left=0pt] {b/$w$(c, b)} (b);
  \draw (c) edge[loop below] node [left=5pt] {c/$w$(c, c)} (c);
\end{tikzpicture}}
  \caption{A bigram graph for the token set \{a,b,c\}.}
  \label{fig:bigram_transitions}
\end{subfigure}
\hspace{2mm}
\begin{subfigure}[b]{0.44\textwidth}
  \centering
  \resizebox{1.0\textwidth}{!}{
    \begin{tikzpicture}
 [>=latex,
  double distance=2pt,
  draw,-latex,
  line width=1pt,
  initial text={},
  shorten >=1pt,
  node distance=3cm,
  auto,
  every state/.style={draw=black, thick}
 ]
  \node[state, initial] (ab) {a, b};
  \node[state, right of=ab] (b) {b};
  \node[state, above right of=b] (ba) {b, a};
  \node[state, right of=b] (bb) {b,b};
  \node[state, below right of=ab] (0) {$\epsilon$};
  \node[state, right of=0] (c) {c};
  \draw (ab) edge[bend left] node {a/$w$(a, b, a)} (ba);
  \draw (ab) edge node {$\epsilon$/$\beta$(a, b)} (b);
  \draw (b) edge node {b/$w$(b, b)} (bb);
  \draw (b) edge node {a/$w$(b, a)} (ba);
  \draw (b) edge[bend right] node [near end, left=5pt] {$\epsilon$/$\beta$(b)} (0);
  \draw (0) edge node {c/$w$(c)} (c);
  \draw (0) edge[bend right] node [near end, right=6pt] {b/$w$(b)} (b);
\end{tikzpicture}}
  \caption{An example of back-off in a trigram graph.}
  \label{fig:backoff_transitions}
\end{subfigure}
  \caption{Transition graph examples. The $n$-gram score is $w$ and $\beta$ is
  the back-off weight.}
\end{figure}

The ASG criterion was constructed with a bigram transition graph $\gB$. We can
use $\gB$ to add transitions to CTC with the same operations in
\eqref{eq:asg_from_graphs} but with the addition of the blank token.

Dense $n$-gram graphs require $O(C^{n-1})$ states and $O(C^n)$ arcs for $C$
tokens, computationally intractable as $n$ and $C$ grow. The dense graph also
suffers from sample efficiency problems as most transitions will not be
observed in the training set. We can alleviate these issues with pruning and
back-off~\citep{katz1987} which can be encoded in a
WFST~\citep{mohri2008speech}. Figure~\ref{fig:backoff_transitions} demonstrates
back-off for a trigram model. The trigram (a, b, a) exists so no back-off is
needed. Neither the trigram (a, b, c) nor the bigram (b, c) exist, so the model
backs off to the $0$-gram state ($\epsilon$) and then transitions to the
unigram state (c).

\subsection{Marginalized Word Piece Decompositions}

Using word pieces has several benefits over grapheme outputs without the sample
and computational efficiency issues, and lexical constraints of outputting
words directly. However, the set of pieces and the decomposition for a word are
learned in an unsupervised and task independent manner~\citep{kudo2018subword,
sennrich2016}. These are then fixed and used to train the task specific model.
The word piece decomposition for a given phrase is not important, serving only
as a stepping stone to more accurate models. This assumption can be made
explicit by marginalizing over the set of decompositions for a target label
while training the task specific model. This allows the model to select the
decomposition(s) which are most salient to the task at hand.

\begin{figure}
\begin{minipage}[b]{0.48\textwidth}
\centering
\begin{subfigure}[b]{1.0\textwidth}
    \centering
    \includegraphics[width=0.8\linewidth]{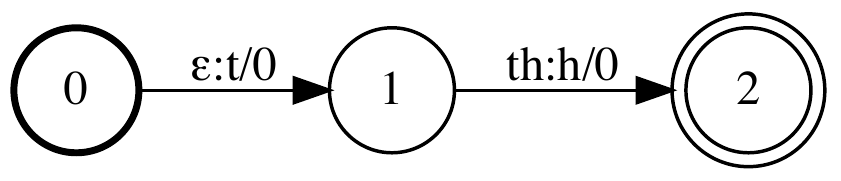}
    \caption{}
    \label{fig:th_wp_to_g}
\end{subfigure}
\begin{subfigure}[b]{1.0\textwidth}
  \centering
    \includegraphics[width=0.9\linewidth]{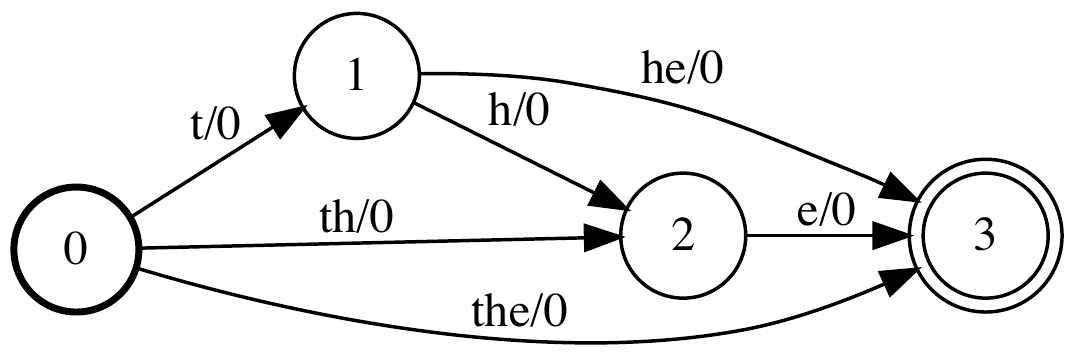}
    \caption{}
    \label{fig:the_decomps}
\end{subfigure}
    \caption{An individual sub-word-to-grapheme transducer (a) for the token
    ``th'' used to construct a lexicon $\gL$ which is used to make the
    decomposition graph (b) for the label ``the''.}
    \label{fig:wp_decomps}
\end{minipage}
  \hspace{2mm}
\begin{minipage}[b]{0.48\textwidth}
    \centering
    \includegraphics[width=1.0\textwidth]{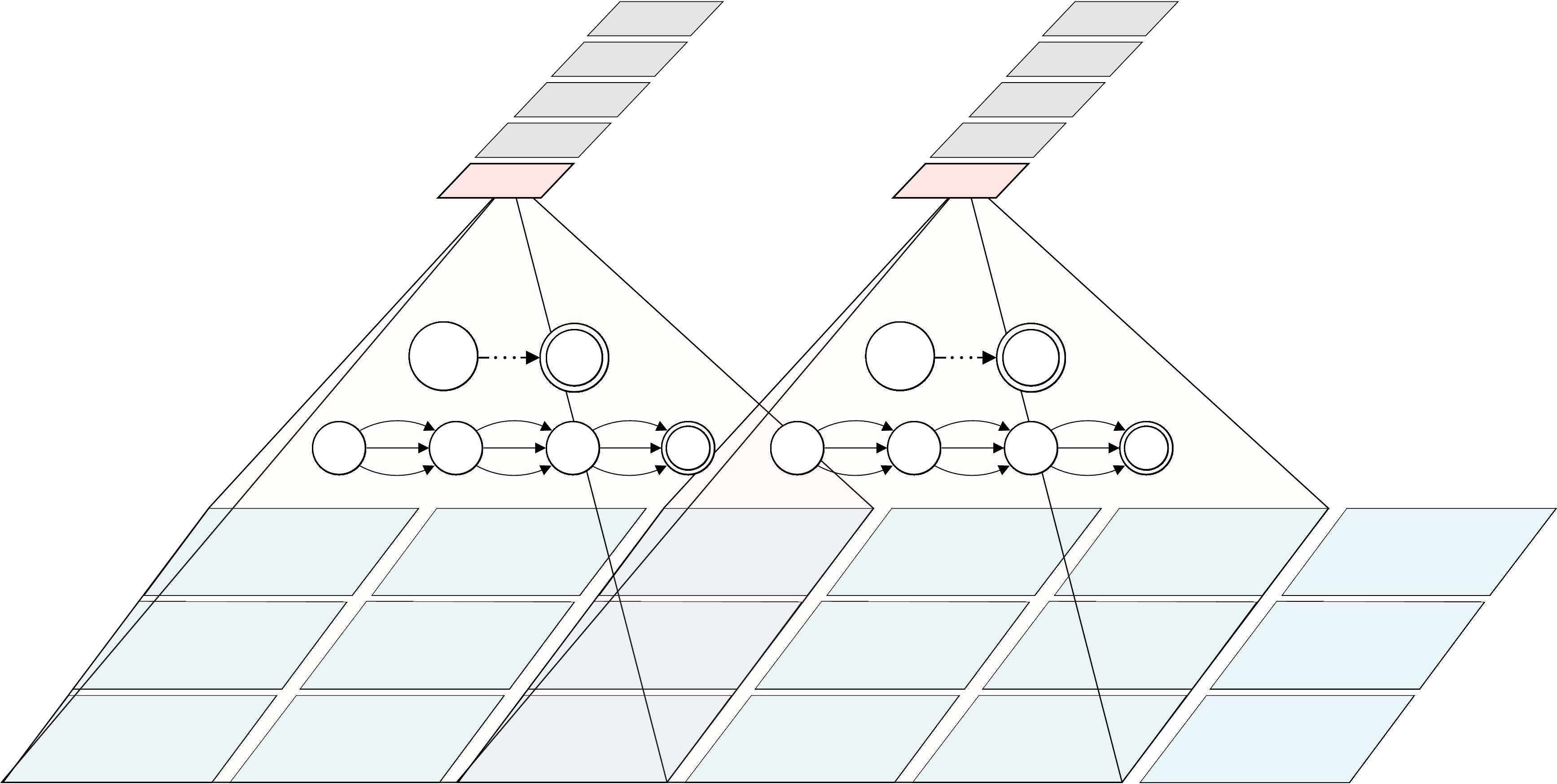}
    \vspace{4mm}
    \caption{The convolutional transducer with a receptive field size of
    $3$ and a stride of $2$. Each output is the forward score of the
    composition of a kernel graph with a receptive field graph.}
    \label{fig:conv_transducer}
\end{minipage}
\end{figure}

Implementing marginalization over word piece decompositions usually requires
non-trivial changes to carefully optimized implementations of existing sequence
criterion. In the differentiable WFST framework this can be implemented in a
plug-and-play fashion by incorporating a single lexicon graph. The lexicon
transducer $\gL$, which maps sequences of sub-word tokens to graphemes, is the
closure of the union of the individual sub-word-to-grapheme graphs. A
composition with the label graph, $\gL \circ \gY$, gives the decomposition
graph for the label $\vy$.  Figure~\ref{fig:wp_decomps} shows an example for
the label ``the''. The decomposition graph can be used in place of the label
graph to construct any down stream sequence criterion. For example, the
constrained graph of
\eqref{eq:asg_from_graphs} becomes:
\begin{equation}
    \label{eq:wp_decomps}
    \gA = \gE \circ (\gB \circ ((\gT_1 + \ldots + \gT_C)^* \circ (\gL \circ \gY)))
\end{equation}
where $\gL = (\gL_1 + \ldots + \gL_C)^*$ and $C$ is the number of word piece tokens.

\subsection{Convolutional WFSTs}

Thus far we have only discussed applications of WFSTs to
combine the output of other models. Figure~\ref{fig:conv_transducer}
demonstrates a convolutional WFST layer which can function as an arbitrary
layer in a tensor-based architecture. Like a standard convolution, the WFST
convolution is specified by a set of kernels, a receptive field size $k$, and a
stride $s$. Each kernel, $\gK_i$, is a WFST. Let the hidden units in a
receptive field at position $t$ be given by $\emH_{t:t+k}$. The hidden units
are the edge weights in the receptive field graph, $\gR_{\emH_{t:t+k}}$. The
output of the $i$-th kernel applied to position $t$ is
\begin{equation}
    \label{eq:conv_trans}
    \emH'_{i, t} = \logadd_{\vp \in \gK_i \circ \gR_{\emH_{t:t+k}}} s(\vp),
\end{equation}
the forward score of the receptive field graph composed with the kernel graph.

The kernel transducers can be structured to impose a desired correspondence.
For example, we construct kernels which map lower-level tokens such as letters
to higher level tokens like word pieces. Without any other constraints, the
interpretation of the input as letter scores and the output as word piece
scores is implicit in the structure of the graphs, and can be learned by the
model up to any indistinguishable permutation. The kernel graphs themselves
have edge weights which, since \eqref{eq:conv_trans} is differentiable, are
easily learned along with the other parameters of the model.

\section{Experiments}
\label{sec:experiments}

We study several of the algorithms from \secref{sec:algorithms} with
experiments in offline handwriting recognition (HWR) and automatic speech
recognition (ASR). A detailed description of the network architectures,
optimization procedures and data processing used is in
Appendix~\ref{sec_apx:experimental_setup}. We report character error rate (CER)
without the use of an external language model or lexical constraint.  Code to
reproduce our experiments is open-source and available at
\url{https://www.anonymized.com}.

\paragraph{Offline Handwriting Recognition} To facilitate comparisons to prior
academic studies, we test our approach on handwritten line recognition using
the IAM database~\citep{marti2002iam}. We use the standard training set and
report results on the first validation set available with the data set
distribution. The splits have changed from those used by prior
work~\citep{graves2008novel, pham2014dropout, voigtlaender2016} which makes
comparisons to such work less meaningful.

\paragraph{Automatic Speech Recognition} Speech recognition experiments  are
performed on the LibriSpeech~\citep{panayotov2015librispeech} and Wall Street
Journal (WSJ)~\citep{paul1992design} corpora. For LibriSpeech, we train with
the ``train-clean-100" subset and report results on the clean validation set.
For WSJ we train on the full corpus and report results on the ``dev-93"
validation set.

\subsection{Transitions}
\label{sec:experiments_ngram}

Table~\ref{tab:hwr_asg_ctc_ngram} shows that including transitions can improve
CER substantially, particularly when no blank token is used. We speculate that
bigram transitions help in this case primarily because of the duration model
they provide. As previously observed~\citep{graves2006}, with a blank token,
the emissions of non-blank outputs spike at a single frame, and no duration
model is needed. In Table~\ref{tab:hwr_pruning}, we see that pruning with
back-off results in no discernible loss in accuracy over a dense transition
model while yielding substantial run-time improvements. With 1000 word pieces,
the epoch time using a graph with pruned bigrams is nearly two orders of
magnitude faster than that of the dense graph.

\begin{table}
\begin{minipage}[b]{0.43\textwidth}
  \small
  \centering
  \caption{A comparison of dense transition graphs from $n=0$ (no
  transitions) up to $n=2$ (bi-grams). We report CER on the IAM validation
  set using letter tokens.}
  \centering
  \begin{tabular}{l l l l}
  \toprule
  \multirow{2}{*}{$n$-gram} & No & Forced  & Optional \\
  & blank & blank & blank \\
  \midrule
  0 & 18.2 & 8.9 & 6.4 \\
  1 & 17.7 & 8.5 & 6.3 \\
  2 & 9.4  & 6.7 & 6.4 \\
  \bottomrule
  \end{tabular}
  \label{tab:hwr_asg_ctc_ngram}
\end{minipage}
\hspace{4mm}
\begin{minipage}[b]{0.52\textwidth}
  \small
  \centering
  \caption{A comparison of bi-gram transition graphs with back-off and varying
  levels of pruning for letters and $1,000$ word pieces. We report CER on the IAM
  validation set and epoch time in seconds.}
  \centering
  \begin{tabular}{l l r l r}
  \toprule
  \multirow{2}{*}{Pruning} & \multicolumn{2}{c}{Letters} & \multicolumn{2}{c}{Word Pieces} \\
    & CER & Time (s) & CER & Time (s) \\
  \midrule
    None & 6.4 & 544 & N/A  & 17,939 \\
    0    & 6.8 & 249 & 22.2 & 683  \\
    10   & 6.5 & 202 & 19.8 & 204  \\
  \bottomrule
  \end{tabular}
  \label{tab:hwr_pruning}
\end{minipage}
\end{table}

\subsection{Marginalized Word Piece Decompositions}

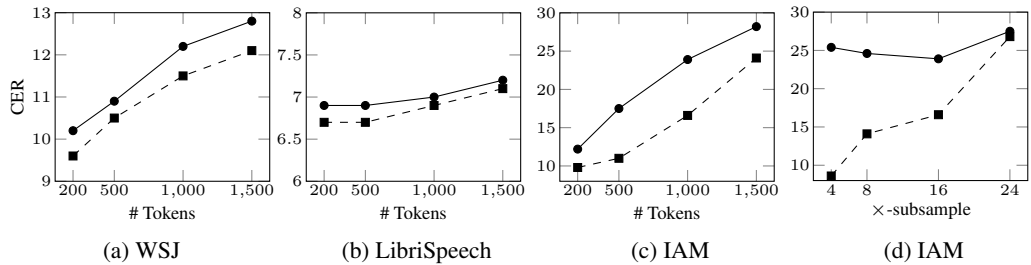
\begin{figure}
\centering
\begin{subfigure}[b]{0.276\textwidth}
  \centering
  \begin{tikzpicture}[trim axis right]
\begin{axis}[
  inner sep=1.5,
  outer sep=0,
  xlabel={\scriptsize \# Tokens},
  ylabel={\scriptsize CER},
  ymin=9,
  ymax=13,
  xtick=data,
  width=1.15\linewidth,
  ylabel near ticks,
  xlabel near ticks,
  ticklabel style={font=\tiny},
%  tick style = {draw=none},
]
\addplot table [y=sp_best, x=num_tokens]{data/wsj_mwpd.dat};
\addplot table [y=mwpd, x=num_tokens]{data/wsj_mwpd.dat};
\end{axis}
\end{tikzpicture}
  \subcaption{WSJ}
  \label{fig:wsj_mwpd_tokens}
\end{subfigure}
\begin{subfigure}[b]{0.235\textwidth}
  \centering
  \begin{tikzpicture}[trim axis left, trim axis right]
\begin{axis}[
  inner sep=1.5,
  outer sep=0,
  xlabel={\scriptsize \# Tokens},
  ymin=6,
  ymax=8,
  xtick=data,
  width=1.35\linewidth,
  ylabel near ticks,
  xlabel near ticks,
  ticklabel style={font=\tiny},
]
\addplot table [y=sp_best, x=num_tokens]{data/librispeech_mwpd.dat};
\addplot table [y=mwpd, x=num_tokens]{data/librispeech_mwpd.dat};
\end{axis}
\end{tikzpicture}
  \subcaption{LibriSpeech}
  \label{fig:librispeech_mwpd_tokens}
\end{subfigure}
\begin{subfigure}[b]{.235\textwidth}
  \centering
  \begin{tikzpicture}[trim axis left, trim axis right]
\begin{axis}[
  inner sep=1.5,
  outer sep=0,
  xlabel={\scriptsize \# Tokens},
  ymin=8,
  ymax=30,
  xtick=data,
  width=1.35\textwidth,
  ylabel near ticks,
  xlabel near ticks,
  ticklabel style={font=\tiny},
]
\addplot table [y=sp_best, x=num_tokens]{data/hwr_mwpd_tokens.dat};
\addplot table [y=mwpd, x=num_tokens]{data/hwr_mwpd_tokens.dat};
\end{axis}
\end{tikzpicture}
  \subcaption{IAM}
  \label{fig:iam_mwpd_tokens}
\end{subfigure}
\begin{subfigure}[b]{0.235\textwidth}
  \centering
  \begin{tikzpicture}[trim axis left, trim axis right]
\begin{axis}[
  xlabel={\scriptsize $\times$-subsample},
  inner sep=1.5,
  outer sep=0,
  ymin=8,
  ymax=30,
  xtick=data,
  width=1.35\linewidth,
  ylabel near ticks,
  xlabel near ticks,
  ticklabel style={font=\tiny},
]
\addplot table [y=sp_best, x=subsample]{data/hwr_mwpd_subsample.dat};
\addplot table [y=mwpd, x=subsample]{data/hwr_mwpd_subsample.dat};
\end{axis}
\end{tikzpicture}
  \subcaption{IAM}
  \label{fig:iam_mwpd_subsample}
\end{subfigure}
  \caption{Validation CER with (dashed) and without (solid) marginalization as
    a function of (a-c) the number of word pieces and (d) the overall sub-sample
    factor using 1000 word pieces.}
  \label{fig:mwpd}
\end{figure}

We experiment with marginalization over word piece decompositions in both ASR
and HWR. We use the SentencePiece toolkit~\citep{kudo2018sentencepiece} to
compute the token set from the training set text. For all datasets, larger
token set sizes result in worse CER, though the effect is mild on LibriSpeech
and more pronounced on IAM. Appendix \figref{fig:wp_counts} suggests this is
primarily due to the different dataset sizes.
Figures~\ref{fig:wsj_mwpd_tokens},\ref{fig:librispeech_mwpd_tokens}, and
\ref{fig:iam_mwpd_tokens} demonstrate that word piece marginalization recovers
some of the accuracy lost when using larger token sets. We examine the Viterbi
decomposition on the LibriSpeech training set. Appendix
table~\ref{tab:viterbi_libri_mwpd} shows that in some cases the most frequently
selected decomposition for a word is more phonologically plausible than the
SentencePiece decomposition.

\begin{figure}
\centering
\begin{subfigure}[b]{0.51\textwidth}
  \centering
  \includegraphics[height=14.5mm]{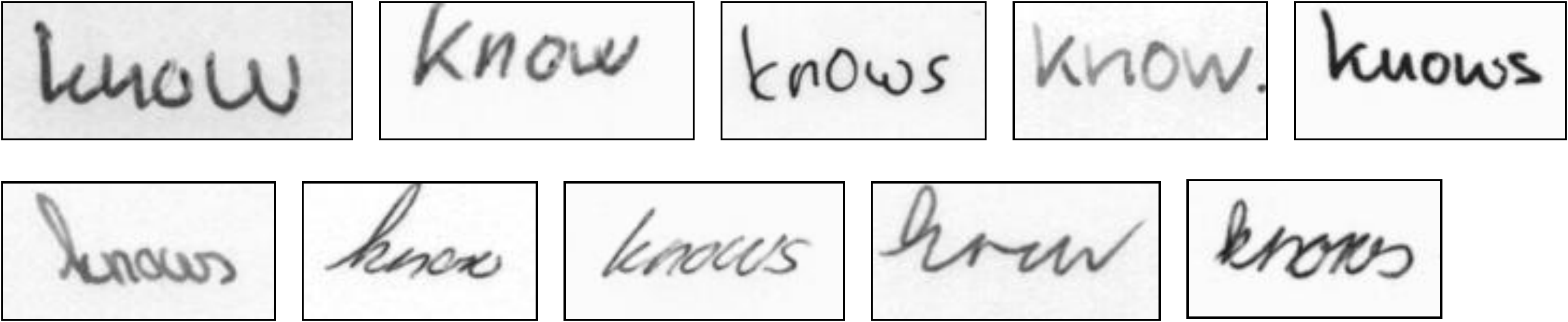}
  \caption{Using word pieces (top) \texttt{\_k,\;n,\;o,\:w} and (bottom) \texttt{\_know}.}
\end{subfigure}
\hspace{3.5mm}
\begin{subfigure}[b]{0.45\textwidth}
  \centering
  \includegraphics[height=14.5mm]{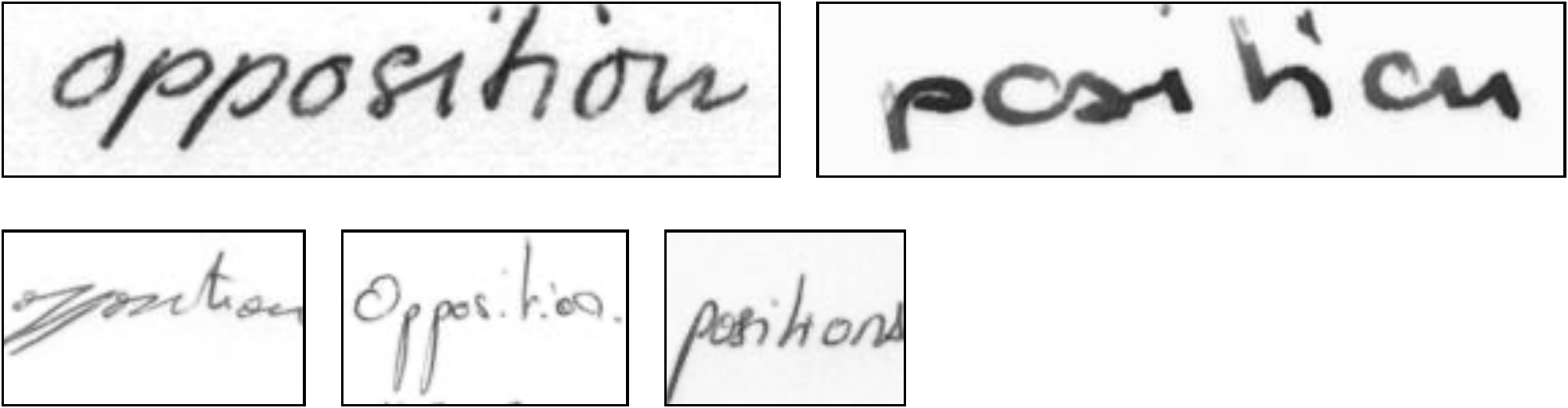}
  \caption{Using word pieces (top) \texttt{p,\,o,\,s,\,i,\,t,\,ion} and
  \allowbreak (bottom) \texttt{position}.}
\end{subfigure}
\caption{The Viterbi word piece decomposition with 1000 tokens for (a)
  ``\_know'' and (b) ``position''. The decompositions are computed on the
  training set images prior to cropping the words.}
\label{fig:word_piece_decomps}
\end{figure}

For IAM, we find that the model favors decompositions using lower-level output
tokens but does still rely on higher-level word pieces in many cases. We also
explore the interaction between sub-sampling in the network and
marginalization. One benefit of marginalization is the ability to dynamically
choose the decomposition suitable to a given frame rate as demonstrated
by~\ref{fig:iam_mwpd_subsample}. We examine the Viterbi decomposition for a few
examples from the training set in~\figref{fig:word_piece_decomps}. We see that
a single model dynamically switches between different decompositions depending
on the input. If the handwriting is tightly spaced, for example, a higher-level
token is preferred.

\subsection{Convolutional WFSTs}

\begin{wraptable}{l}{6.5cm}
    \captionof{table}{A comparison of the convolutional WFST layer to a traditional
    convolution. We report the CER on the IAM validation set and compare the
    two layers in number of parameters and number of operations.}
    \centering
    \small
    \begin{tabular}{l c c r c}
    \toprule
    Convolution & CER & Params & Ops \\
    \midrule
    WFST        & 19.5 & 2,048 & $O(k w c_o)$ \\
    Traditional & 20.7 & 79,000 & $O(k c_i c_o)$ \\
    \bottomrule
    \end{tabular}
    \label{tab:conv_wfst}
\end{wraptable}

We experiment with the convolutional WFST layer on the IAM data set. The
kernels in the WFST layer transduce letters to 200 word pieces. We use a
CTC-style graph structure (fig.~\ref{fig:ctc_alignments}) for the kernels and
hence the input to the WFST has an additional dimension for the
\ctcblank\,token. We place the WFST convolution in between the third and fourth
TDS groups. The model is trained to predict 1000 word pieces at the output
layer. We compare the WFST convolution with a traditional convolution in the
same position. Both convolutional layers have a kernel width $k=5$, a stride of
$4$, $c_i=80$ input channels, and $c_o=200$ output channels.
Table~\ref{tab:conv_wfst} demonstrates that the WFST convolution outperforms
the traditional convolution. The WFST convolution is also less expensive as the
number of parameters and the number of operations needed scale with the lengths
of the word pieces in letters ($w$ in table~\ref{tab:conv_wfst}) rather than
the number of input channels.

\section{Conclusion}
\label{sec:conclusion}

We have shown that combining automatic differentiation with WFSTs is not only
possible, but a promising route to new learning algorithms and architectures.
The design space of structured loss functions that may be encoded and
efficiently implemented through the use of differentiable WFSTs is vast. We
have only begun to explore it.

We see many other related and exciting paths for future work in automatic
differentiation with WFSTs. First, we aim to continue to bridge the gap between
static computations with WFSTs while performing inference and dynamic
computations with WFSTs during model training. This will require further
optimizations and potentially approximations when integrating higher level
lexical and language modeling constraints. Second, the use of WFSTs as a
substitute for tensor-based layers in a deep architecture is intriguing. This
may be more effective than traditional layers in imposing prior knowledge on
the relationship between higher-level and lower-level representations or in
discovering useful representations from data.

\subsubsection*{Acknowledgments}
Thanks to Sam Gross for help circumventing the Python GIL. Thanks to Ronan
Collobert, Gabriel Synnaeve, Abdelrahman Mohamed and Adrien Dufraux for
helpful discussions and comments.

\bibliography{references}
\bibliographystyle{iclr2021_conference}

\appendix
\section{Algorithms}

\subsection{Repeat Tokens in CTC}
\label{sec_apx:ctc_repeats}

The blank token in CTC (\ctcblank) also serves to disambiguate consecutive
repeat tokens in the output from a single token corresponding to two input
frames. Enforcing this construction in the WFST framework requires keeping
track of the previously generated token and only allowing transitions on a new
token or on $\ctcblank$. Let $\gT_1, \ldots, \gT_C$ be the individual token
graphs, including the blank token graph as in \figref{fig:ctc_blank}. The full
token graph to disallow repeat consecutive transitions is given by
\begin{equation}
 \label{eq:ctc_no_repeats}
 \gT = \left(\sum_{i=1}^C \sum_{j=1, j \ne i}^C \gT_i\gT_j  \right)^*.
\end{equation}
In practice, constructing the graph with \eqref{eq:ctc_no_repeats} will result
in more states and arcs than are needed. Instead, we construct the graph
directly. However, we note that these operations may be used along with
$\epsilon$ removal and state minimization to keep the resulting token graph
$\gT$ small.

\subsection{Priors with  CTC}
\label{sec_apx:ctc_priors}

The individual token graphs in CTC do not assume much about the duration or
spacing of the tokens at the input level. The graph in \figref{fig:asg_token_a}
states only that non-blank tokens must align to at least one frame. The graph
in ~\figref{fig:ctc_blank} states that the \ctcblank\, token is optional.

\begin{figure}
    \centering
    \begin{subfigure}[b]{0.17\textwidth}
        \includegraphics[width=\textwidth]{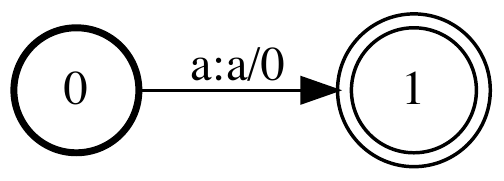}
        \caption{Spike CTC.}
        \label{fig:spike_ctc}
    \end{subfigure}
    \hspace{2mm}
    \begin{subfigure}[b]{0.28\textwidth}
        \includegraphics[width=\textwidth]{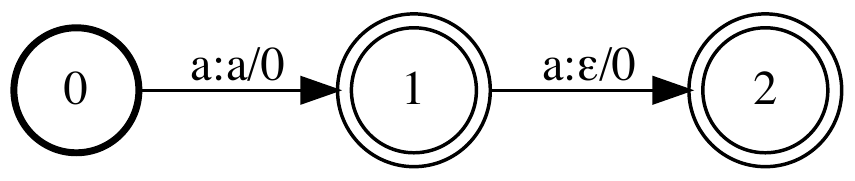}
        \caption{Duration-limited CTC.}
        \label{fig:duration_limited_ctc}
    \end{subfigure}
    \hspace{2mm}
    \begin{subfigure}[b]{0.48\textwidth}
        \includegraphics[width=\textwidth]{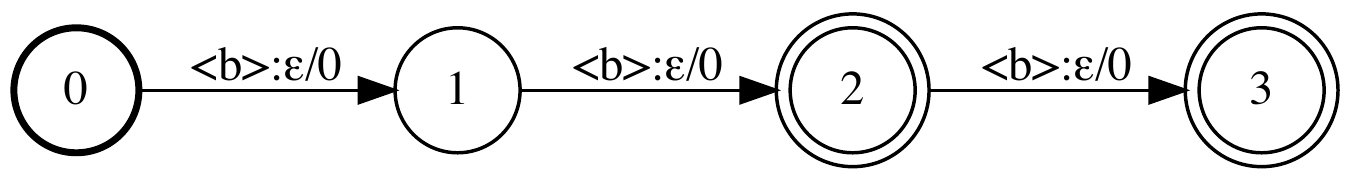}
        \caption{Equally spaced CTC.}
        \label{fig:space_limited_ctc}
    \end{subfigure}
    \caption{A few individual token graphs used to construct variants of, for
    example, CTC. The individual token graphs are combined to create the
    overall token graph $\gT = (\gT_1 + \ldots + \gT_C)^*$.}
\end{figure}

The differentiable WFST framework simplifies the construction of variations of
these token-level graphs with potentially useful alternative assumptions. For
example, we can construct a ``spike'' CTC which only allows a single repetition
of a non-blank label by using the graph in \figref{fig:spike_ctc}.  Similarly,
a ``duration-limited'' CTC could use a token graph as in
\figref{fig:duration_limited_ctc}. In this case, the duration of a token is
limited to be between one and two input frames. We could specify a range of
allowed distances between non-blank tokens using the blank token graph in
\figref{fig:space_limited_ctc}. These may all be mixed and matched on a
per-token basis.

\section{Experimental Setup}
\label{sec_apx:experimental_setup}

In both handwriting recognition and speech recognition experiments, we use a
variation of the time-depth separable (TDS) convolutional
architecture~\citep{hannun2019}. The code and settings needed to reproduce our
experiments are available at \url{https://www.anonymized.com}.

\subsection{Handwriting Recognition}

For experiments in HWR, we propose a variant of the TDS model which preserves
the 2D structure in the image. The 2D TDS block first applies a 3D convolution
with kernel size $1 \times k_h \times k_w$ on a 4-dimensional input of size $c
\times d \times h \times w$ where $c$ and $d$ are the separated channel and
depth dimensions and $w$ and $h$ are the height and width of the image
respectively. Following the 3D convolution, is a $1\times 1$ convolution with
$c \times d$ input and output channels. We apply dropout, the ReLU
non-linearity and residual connections as in the original TDS architecture. In
place of layer normalization, we use 2D instance
normalization~\citep{isola2017image} with a learned affine transformation.

Unless otherwise noted, we use $4$ groups of three TDS blocks each. In between
each TDS block we apply a standard 2D convolution and optionally sub-sample the
image in the height and width dimensions. For letter-based experiments we
sub-sample the height by $2$ and the width by $2$ only after the first
two groups. For word piece experiments we sub-sample both the height and width
by a factor of $2$ after every group for an overall sub-sampling factor of
$16\times$ in each dimension. We use a uniform kernel size of $5 \times 7$. We
start the number of channels ($c\times d$) at $8$ and increase it to $16$, $32$
and $64$ in between each TDS group.

For optimization we use a simple stochastic gradient descent with an initial
learning rate of $10^{-1}$ which is reduced every $100$ epochs by a factor of $2$.
All models are trained for $400$ epochs and we save the model which achieves
the best results on the validation set during training. We use a mini-batch
size of 32 with distributed data parallel training.

Bounding boxes for the line-level segmentation for each page are provided along
with the ground truth texts. We perform no other data pre-processing other than
a simple normalization, removing the global pixel mean and standard deviation
from each image. We use three forms of data augmentation--a random resize and
crop, a random rotation, and a random color change.

\subsection{Speech Recognition}

For experiments in speech recognition, we use a 1D TDS architecture with 3
groups each containing 5 TDS blocks. In between each group we sub-sample the
input in the time-dimension by a factor of 2 for an overall sub-sampling factor
of 8. The TDS blocks all use a kernel width of $5$. The first group has $4$
channels, and we double the number of channels in between each group.

For optimization we use stochastic gradient descent. Initial learning rates are
tuned for each model type in the range $10^{-2}$ to $1$. The learning rate is
reduced every $100$ epochs by a factor of $2$.  All models are trained for
$400$ epochs and we save the model which achieves the best results on the
validation set during training. We use a mini-batch size of 16 with distributed
data parallel training.

For features, we use 80 log-mel-scale filter banks on windows of 20
milliseconds every 10 milliseconds. We normalize individual examples by
removing the mean and dividing by the standard deviation. We apply SpecAugment
for regularization following the ``LD'' policy but without
time-warping~\citep{park2019specaugment}.

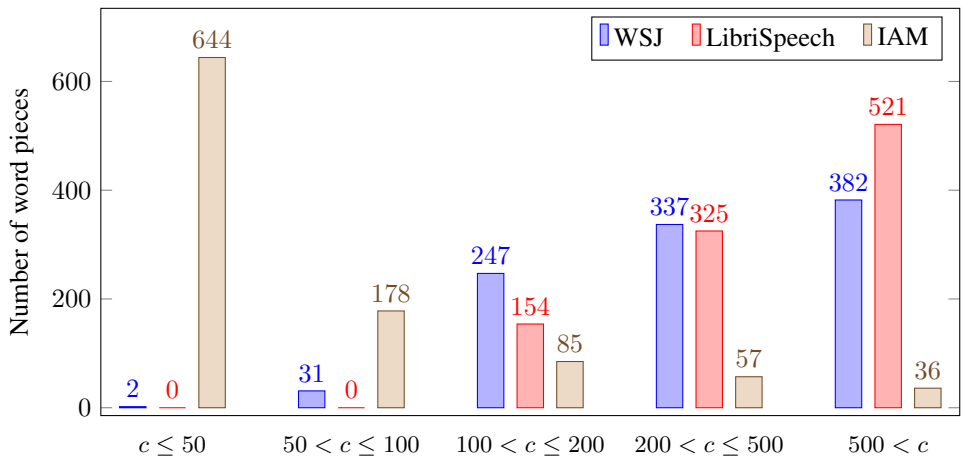
\begin{figure}
  \centering
  %% Only show a single bar in legend
\pgfplotsset{compat=1.11,
    /pgfplots/ybar legend/.style={
    /pgfplots/legend image code/.code={%
       \draw[##1,/tikz/.cd,yshift=-0.25em]
        (0cm,0cm) rectangle (3pt,0.8em);},
   },
}
\begin{tikzpicture}
\begin{axis}[
    width = 13cm,
    height = 7cm,
    ybar = 5pt,
    enlarge y limits = 0.02,
    enlarge x limits = 0.1,
    ymax = 720,
    legend style={
      legend columns=-1,
      /tikz/every even column/.append style={column sep=0.3cm}
    },
    ylabel={Number of word pieces},
    symbolic x coords={
      $c \le 50$, $50 < c \le 100$,
      $100 < c \le 200$, $200 < c \le 500$, $500 < c$
    },
    xtick=data,
    nodes near coords,
    nodes near coords align={vertical},
    x tick label style = {font=\small},
    xtick style = {draw=none}
]
%% WSJ
\addplot coordinates {
  ($c \le 50$, 2) ($50 < c \le 100$, 31) ($100 < c \le 200$, 247)
  ($200 < c \le 500$, 337) ($500 < c$, 382)
};
%% LibriSpeech 
\addplot coordinates {
  ($c \le 50$, 0) ($50 < c \le 100$, 0) ($100 < c \le 200$, 154)
  ($200 < c \le 500$, 325) ($500 < c$, 521)
};
%% IAM
\addplot coordinates {
  ($c \le 50$, 644) ($50 < c \le 100$, 178) ($100 < c \le 200$, 85)
  ($200 < c \le 500$, 57)  ($500 < c$, 36)
};
\legend{WSJ, LibriSpeech, IAM}
\end{axis}
\end{tikzpicture}
  \caption{The number of word pieces with occurrences $c$ in the given range.
  For each dataset we use 1,000 word piece tokens and count the number of
  occurrences in the training text using the decomposition for each word
  provided by SentencePiece. The WSJ training text contains 639k words, the
  LibriSpeech training text has 990K words and the IAM training text has only
  54K words.}
  \label{fig:wp_counts}
\end{figure}

\begin{table}
  \centering
  \caption{A comparison of the most frequent decomposition to the SentencePiece
  decomposition for a given word. The words were selected by first computing
  the Viterbi decomposition for each example in the LibriSpeech training set.
  We then chose words for which the most frequently selected decomposition was
  different from the SentencePiece decomposition. The counts are the
  occurrences of each decomposition in the Viterbi paths on the training set.
  In some cases the most frequent decomposition appears to be more
  phonologically plausible than the SentencePiece decomposition.}
  \label{tab:viterbi_libri_mwpd}
  \begin{tabular}{l l l l l}
    \toprule
    \multirow{2}{*}{Word} & \multicolumn{2}{c}{Most common} & \multicolumn{2}{c}{SentencePiece} \\
     & Decomposition & Count & Decomposition & Count \\
    \midrule
    able & \texttt{\wordsep{a}, ble} & 121 & \texttt{\wordsep{}, able} & 88 \\
    wind & \texttt{\wordsep{w}, in, d} & 90 & \texttt{\wordsep{wi}, n, d} & 72 \\
    single & \texttt{\wordsep{si}, ng, le} & 65 & \texttt{\wordsep{}, s, ing, le} & 40 \\
    move & \texttt{\wordsep{mo}, ve} & 54 & \texttt{\wordsep{mov}, e} & 47 \\
    ring & \texttt{\wordsep{ri}, ng} & 48 & \texttt{\wordsep{r}, ing} & 23 \\
    \bottomrule
  \end{tabular}
\end{table}

\end{document}